\documentclass[times,review,10pt]{elsarticle}

\usepackage{amssymb}
\usepackage{amsmath}
\usepackage{lineno}
\usepackage{hyperref}
\usepackage{cleveref}
\usepackage{tcolorbox}
\usepackage{multirow}
\usepackage{booktabs}

\journal{Pattern Recognition}

\begin{document}
\begin{frontmatter}



\title{SceneVTG++: Controllable Multilingual Visual Text Generation in the Wild}

\author[hust]{Jiawei Liu\corref{sign}}\ead{jiaweiliu@hust.edu.cn}
\author[alibaba]{Yuanzhi Zhu}\ead{zyz.kpillow@gmail.com}
\author[alibaba]{Feiyu Gao}\ead{feiyu.gfy@alibaba-inc.com}
\author[alibaba]{Zhibo Yang}\ead{yangzhibo450@gmail.com}
\author[alibaba]{Peng Wang}\ead{wdp0072012@gmail.com}
\author[alibaba]{Junyang Lin}\ead{junyang.ljy@alibaba-inc.com}
\author[hust]{Xinggang Wang}\ead{xgwang@hust.edu.cn}
\author[hust]{Wenyu Liu\corref{mycorrespondingauthor}}\ead{liuwy@hust.edu.cn}

\cortext[mycorrespondingauthor]{Corresponding author}
\cortext[sign]{This work was done when Jiawei Liu were interns at Alibaba Group.}

\address[hust]{School of EIC, Huazhong University of Science and Technology, Wuhan, 430074, China}
\address[alibaba]{Alibaba Group, Hangzhou, 311121, China}




\begin{abstract}
Generating visual text in natural scene images is a challenging task with many unsolved problems. 
Different from generating text on artificially designed images (such as posters, covers, cartoons, etc.), the text in natural scene images needs to meet the following four key criteria: 
(1) Fidelity: the generated text should appear as realistic as a photograph and be completely accurate, with no errors in any of the strokes.
(2) Reasonability: the text should be generated on reasonable carrier areas (such as boards, signs, walls, etc.), and the generated text content should also be relevant to the scene.
(3) Utility: the generated text can facilitate to the training of natural scene OCR (Optical Character Recognition) tasks.
(4) Controllability: The attribute of the text (such as font and color) should be controllable as needed.
In this paper, we propose a two stage method, SceneVTG++, which simultaneously satisfies the four aspects mentioned above.
SceneVTG++ consists of a Text Layout and Content Generator (TLCG) and a Controllable Local Text Diffusion (CLTD). 
The former utilizes the world knowledge of multi modal large language models to find reasonable text areas and recommend text content according to the nature scene background images, while the latter generates controllable multilingual text based on the diffusion model.
Through extensive experiments, we respectively verified the effectiveness of TLCG and CLTD, and demonstrated the state-of-the-art text generation performance of SceneVTG++. 
In addition, the generated images have superior utility in OCR tasks like text detection and text recognition. 
Codes and datasets will be available.
\end{abstract}



\begin{keyword}
Visual text generation \sep Real-world scenarios \sep Conditional diffusion models



\end{keyword}

\end{frontmatter}



\section{Introduction}
\label{sec1}
In recent years, diffusion models have sparked intense research interest in the field of generative AI. 
Remarkable works like DALLE~\cite{DBLP:conf/icml/RameshPGGVRCS21,ramesh2022hierarchical,BetkerImprovingIG} and Stable Diffusion~\cite{DBLP:conf/cvpr/RombachBLEO22,DBLP:conf/iclr/PodellELBDMPR24} have achieved outstanding progress in the field of text-to-image generation. 
Their remarkable image generation results have captured the intense attention of both academia and industry.
However, these methods still have significant deficiencies when generating images with text~\cite{DBLP:journals/corr/abs-2206-00169,DBLP:conf/acl/LiuGSCRNBM0C23}, especially when generating natural scene visual text, which hinders their prospects for real-world applications.
When it comes to visual-text-related research, scene text is considered to be one of the most difficult research subjects, with a major obstacle being the scarcity of sufficient and diverse data resources.
This is due to the fact that real-world scene text encompasses a wide range of complex backgrounds, font styles, font colors, languages, and so on. 
Collecting and annotating large-scale real-world scene texts requires a huge amount of human and financial resources.
Therefore, synthetic data stands out as a crucial data source for training tasks involving natural scene text.
Extensive studies have already demonstrated the significance of synthetic data in scene text-related tasks~\cite{DBLP:conf/cvpr/FangXWM021, DBLP:conf/cvpr/LiuCSHJW20, DBLP:journals/pami/ShiBY17, DBLP:conf/cvpr/ZhouYWWZHL17}, making scene text generation a popular research topic. 
Existing methods for visual text generation are primarily divided into two categories, one of which is based on rendering engines to generate text on background images, mostly utilized for training OCR tasks~\cite{DBLP:conf/cvpr/GuptaVZ16, DBLP:journals/corr/abs-2003-10608}.
The other type of visual text generation method based on diffusion models~\cite{DBLP:conf/icml/NicholD21, DBLP:conf/iclr/SongME21} has been extensively explored by many researchers, and significant progress has been made~\cite{DBLP:conf/nips/ChenHL0CW23, DBLP:journals/corr/abs-2311-16465, DBLP:conf/iclr/TuoXHGX24}. 
However, there is still a lack of research on natural scene text generation.
We have summarized the issues encountered by these two methods when generating scene text, and believe that the generated text should meet four key criteria illustrated in \Cref{fig:1}.
The detailed descriptions of these four criteria are as follows:

\begin{figure*}
  \centering
  \includegraphics[width=0.98\linewidth]{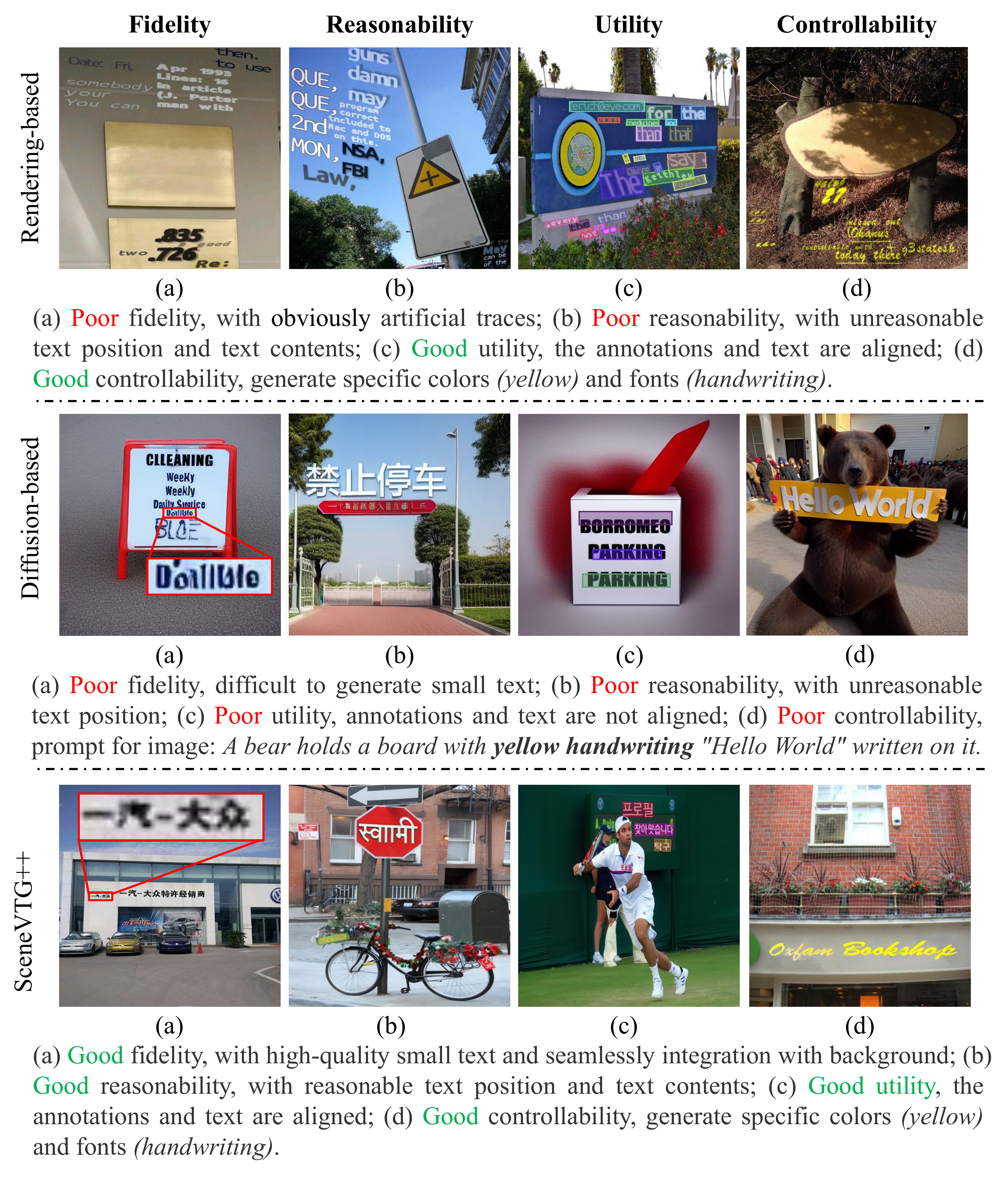}
  \caption{In terms of fidelity, reasonability, utility, and controllability, our proposed SceneVTG++ excels in adequately satisfying all four criteria when compared with various other methods. Zoom in for better views.}
  \label{fig:1}
\end{figure*}

\begin{itemize}
\item \textbf{Fidelity}: Text image pixels should seamlessly blend with the background, with no visible artifacts. And there should be no incorrect strokes in the text.

\item \textbf{Reasonability}: The text should be generated in an appropriate position and have an aesthetically pleasing layout. The content of the text should be relevant to the context of the image rather than randomly selected from predefined corpus. 

\item \textbf{Utility}: The layout and content of the generated text should match the intended exactly, so as to enhance the performance of relevant tasks using the generated images, such as OCR detection and 
OCR recognition.

\item \textbf{Controllability}: The generated text should be controllable, allowing people to freely modify the attributes of the text as needed.
\end{itemize}

\begin{figure*}
  \centering
  \includegraphics[width=0.98\linewidth]{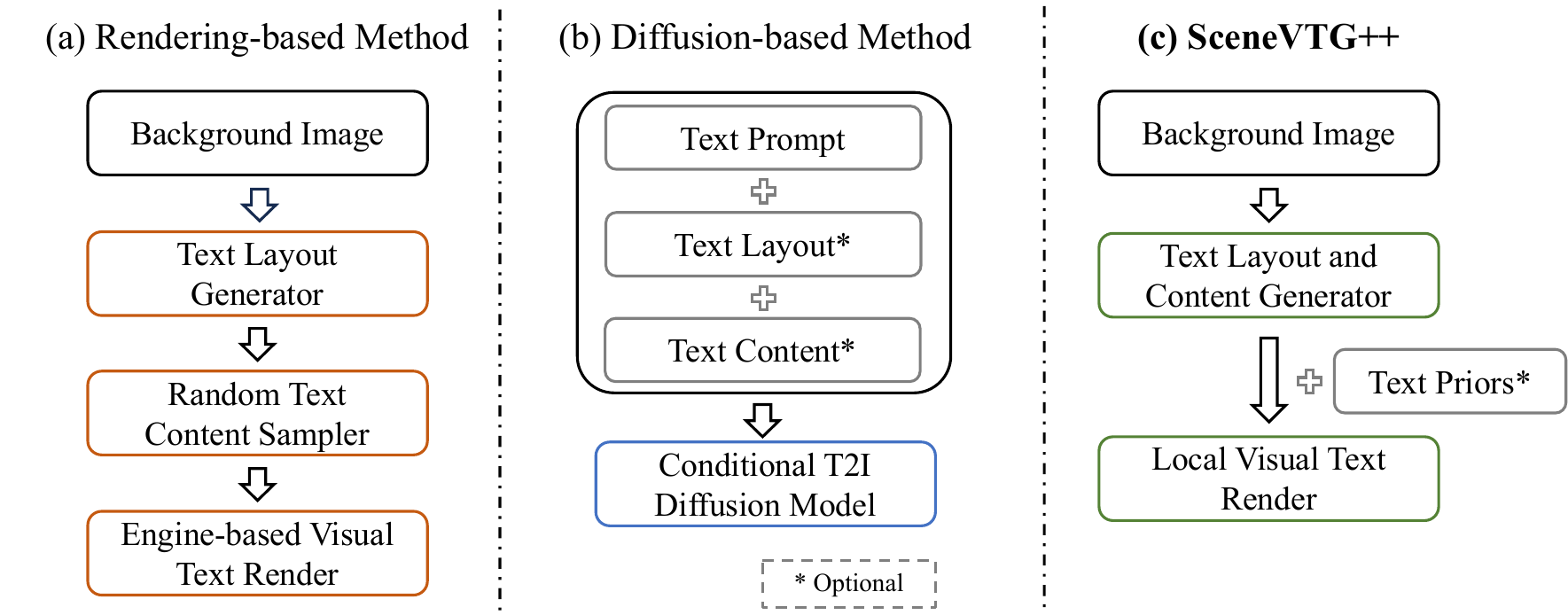}
  \caption{Comparison of pipelines for different visual text generation methods}
  \label{fig:2}
\end{figure*}

\Cref{fig:2} (a) shows the pipeline of rendering-based method, which consists of three steps: identifying suitable text layouts via various feature maps (e.g. segmentation maps, edge maps, etc.); randomly choosing text contents from the corpus; and applying manually defined render engines to generate text across random materials (e.g., colors, fonts, and so on) .
As shown in~\Cref{fig:1} (a), visual text images generated by this method perform well in the aspects of utility and controllability, making them effective for training OCR tasks such as OCR detection and OCR recognition.
However, due to the following issues, they fall short in fidelity and reasonability: (1) manually crafted generation rules struggle to adapt to all backgrounds, leading to noticeable artificial traces; (2) the determined text layouts may be located in unreasonable positions such as the sky or the ground, and the randomly selected text content is also semantically unrelated to the image background, leading to unreasonable text generation in the image.
These issues result in considerable differences between the distribution of the generated data and real-world data.

\Cref{fig:2} (b) shows the pipeline of the diffusion-based methods, which typically involves generating visual text with background images according to the given prompts, and incorporating additional text layouts, contents, etc.
\Cref{fig:1} (b) illustrates that the visual text generated by this methods falls short in terms of fidelity, reasonability, utility, and controllability.
The main issues are as follows: 
(1) Existing methods encounter difficulties in generating small characters. 
(2) The reasonability of generating text layout and content depends on the given prompts, but a rational prompt requires careful design manually. 
(3) Existing methods cannot precisely generate text at the given layout location and may generate unexpected text.
(4) The generated text are often uncontrollable, leading to significant differences in text styles for the same image.
These problems have led to the fact that, although the integration of foreground texts and background images is more seamless than in rendering-based methods, it has not yet been used for synthesizing data for OCR tasks.

In this paper, we propose a framework called SceneVTG++, which further extends the fidelity, reasonability, utility, and controllability of generating visual text while retaining the advantages of the diffusion-based methods.
\Cref{fig:2} (c) shows the pipeline of SceneVTG++, which adopts a two-stage paradigm, consisting of the Text Layout and Content Generator (TLCG) and the Controllable Local Text Diffusion (CLTD). 
In TLCG, thanks to the remarkable visual reasoning capabilities of Multi-Modal Large Models (MMLMs), we have simultaneously achieved reasonable text layout generation and multilingual text content recommendations based on the background images.
In CLTD, we leverage a pixel-level conditional diffusion model without latent compression module to achieve text generation at various scales.
The location and content of the text generated by CLTD are based on the output from TLCG, achieving automated scene text generation.
In addition, CLTD introduces the control of text priors to achieve controllable scene text generation.
To achieve our goals, we contribute a real-world dataset called SceneVTG-Erase++.
The dataset contains a total of 197K multi-lingual scene text image pairs, each pair including a version with text and a version without text. 
Additionally, each image pair comes with OCR annotations and textual attribute information such as color and font.
In addition, we create a synthetic dataset SceneVTG-Syn with 100K image pairs, expanding the diversity of text attributes.
This paper extends our preliminary conference paper~\cite{zhu2024visual} by implementing multilingual text generation and attribute control of the generated text.
The main contributions can be outlined as follows::

(1) We propose a novel two-stage framework for visual scene text generation called SceneVTG++ with two new datasets SceneVTG-Erase++ and SceneVTG-Syn. It is capable of synthesizing realistic multilingual scene text images, serving as an expansion of real-world natural scene text data.

(2) In our framework, we develop the Text Layout and Content Generator (TLCG), which leverages the visual reasoning capabilities of Multi-Modal Large Models, and the Controllable Local Text Diffusion
 (CLTD) to enable controllable multilingual scene text generation.

(3) Extensive qualitative and quantitative results demonstrate that the visual text generated by SceneVTG++ are significantly superior to previous methods in four aspects: fidelity, reasonability, utility and controllability.

\section{Related work}
\label{sec2}
\subsection{Text Layout generation}
\subsubsection{Contents-agnostic text layout generation}
Content-agnostic text layout generation methods usually organize layouts on a blank canvas with constraints.
These methodologies employ diverse neural architectures to decipher the underlying principles and rules of layout from training data.
For instance, LayoutGAN~\cite{DBLP:journals/pami/LiYHZX21} harnesses the generative adversarial network (GAN) framework to generate document layouts. 
LayoutVAE~\cite{DBLP:conf/iccv/JyothiDHSM19} adopts the mechanics of a Variational Autoencoder~\cite{DBLP:journals/corr/KingmaW13}, while LayoutTransformer~\cite{DBLP:conf/iccv/GuptaLA0MS21} leverages the powerful autoregressive capabilities of the Transformer~\cite{DBLP:conf/nips/VaswaniSPUJGKP17} architecture. Adding to this ensemble, LayoutDM~\cite{DBLP:conf/cvpr/InoueKSOY23} pioneers the use of a diffusion model grounded in discrete state spaces, further expanding the possibilities and diversity of layout generation.
Despite these methods prowess in crafting visually pleasing layouts that enhance design processes in domains like mobile UIs and document design,  they confront limitations in real scene implementations.
Their utility is hindered as they specialize in populating blank canvases with text, ignoring the context and content within the images.

\subsubsection{Contents-aware text layout generation}
The process of content-aware text layout generation involves carefully organizing and placing text in an image to create the most visually reasonable result, which is a crucial step in synthesizing natural scene text images.
Previous methods mostly depend on utilizing different feature maps to ascertain text layout.
SynthText~\cite{DBLP:conf/cvpr/GuptaVZ16} leverages various feature maps (depth estimation maps, color and texture segmentation maps) to identify optimal text layouts.
VISD generates layouts by referencing the semantics of objects in the image, ensuring that the layouts fall entirely within a semantically consistent area.
UnrealText~\cite{long2020unrealtext} uses the 3D information of images and detects suitable text layouts based on object meshes.
LBTS~\cite{DBLP:journals/tip/TangMO23} trained a segmentation network based on text-erased images to identify optimal areas for text placement within images.
The drawback of these methods is that they ignore the context of the background images.
For example, these methods might generate text in the sky or on the ground, but this is unreasonable in real-world scenarios.
In addition, the randomly selected text content is unrelated to the image context.

In addition to natural scenes, there is also a lot of work in the field of poster design that focuses on generating content-aware text layouts.
CGL-GAN~\cite{DBLP:conf/ijcai/ZhouXMGJX22} explicitly inputs salience maps as additional information into the designed Transformer-GAN based model to generate text layouts.
Similar work leveraging external saliency maps includes PosterLayout~\cite{DBLP:conf/cvpr/HsuHPKZ23}, which is based on CNN-LSTM.
RADM~\cite{DBLP:conf/cikm/LiLFZLZLZ0LS23} is the first to consider the content of the text when generating text layout, using a diffusion model to generate text boxes of appropriate length based on the length of the text.
Recently, remarkable advancements in text inference capabilities of large language models have attracted widespread attention, some layout generation methods have begun to harness the power of LLMs.
LayoutPrompter~\cite{DBLP:conf/nips/LinGS0L023} uses bounding boxes in nature language to represent the positions of salient objects , and utilizes a large language model to generate text layouts.
PosterLlama~\cite{seol2024posterllama} adopts the structure of a multimodal large language model and train an adapter to achieve more precise content-aware text layout generation.
Despite demonstrating commendable performance within poster layout design, these techniques prove inadequate when apply to natural scenarios.
Inspired by the remarkable visual reasoning capabilities of Multi-Modal Large Models (MMLMs), our conference version SceneVTG~\cite{zhu2024visual} achieves reasonable text layouts and contents generation in natural scenes.
We extend the text content generation to multiple languages and improve the performance of layout generation by employing a full fine-tuning training strategy.

\subsection{Image generation}
\subsubsection{Controllable image generation}

The exploration of generating controllable images has long captured the attention of researchers and experts.
Take CGAN~\cite{mirza2014conditional} and CVAE~\cite{DBLP:conf/nips/SohnLY15} as examples, these innovative models are the outcomes of early researchers' quest for the controllability of GAN and VAE.
Recent research has been dedicated to exploring the controllability of diffusion models.
DaLLE-2~\cite{ramesh2022hierarchical} uses pre-trained CLIP~\cite{DBLP:conf/icml/RadfordKHRGASAM21} text encoder to extract text features as conditional guidance for image generation. 
Latent Diffusion~\cite{DBLP:conf/cvpr/RombachBLEO22} incorporates conditional encoding for various formats (such as segmentation maps, text, and images) and applies diffusion models in latent space to reduce the cost of training and inference. 
ControlNet~\cite{DBLP:conf/iccv/ZhangRA23} extends the capabilities of Latent Diffusion by introducing a network structure that allows for the separate integration of different types of conditions into pre-trained diffusion models, eliminating the need to retrain these conditions from scratch.
Composer~\cite{DBLP:conf/icml/HuangC0SZZ23} decomposes images into various attribute factors and then trains diffusion models with all attribute factors as conditions to control image generation in terms of color, style, semantics, and other aspects. 
While these conditional diffusion models excel in generating high-quality images, they struggle with text generation and have not effectively mastered the control of text attributes such as text fonts and text colors.

\subsubsection{Visual text generation}
\label{subsec2.2.2}
We categorize the scene text generation task into two types based on the methods used: full image generation and local text generation.
Full image generation follows the paradigm of text to image generation, simultaneously creating text with background images based on prompts, resulting text generation that seamlessly integrate with the background.
GlyphControl~\cite{DBLP:conf/nips/YangGYLDH023} trained a Glyph ControlNet on the basis of pre-trained Stable Diffusion~\cite{DBLP:conf/cvpr/RombachBLEO22} to control the generated text.
TextDiffuser~\cite{DBLP:conf/nips/ChenHL0CW23} utilizes LayoutTransformer~\cite{DBLP:conf/iccv/GuptaLA0MS21} to generate segmentation masks with character-level information, and then generates text based on these masks.
TextDiffuser-2~\cite{DBLP:journals/corr/abs-2311-16465} improves the method of generating layouts by fine-tuning a large language model to create more diverse text layouts.
Anytext~\cite{DBLP:conf/iclr/TuoXHGX24} organizes and constructs a large-scale multilingual text generation dataset, and for the first time implements multilingual text generation based on ControlNet.
Some concurrent works, such as GlyphByT5~\cite{DBLP:journals/corr/abs-2403-09622}, retrained a text encoder and designed a region-based cross-attention
mechanism to incorporate conditions into the diffusion model, achieving the generation of long texts. 
TextGen~\cite{DBLP:journals/corr/abs-2407-11502} utilized ControlNet to construct a global control stage and a detail control stage, enabling them to be activated at different timesteps respectively.
However, these methods face the issue of limited text generation positions~\cite{DBLP:journals/corr/abs-2311-16465} and poor performance in generating small text~\cite{DBLP:conf/iclr/TuoXHGX24}, making it difficult to apply these methods to natural scene text generation.

Local text generation involves rendering text on a given background image.
Utilizing GAN, some methods~\cite{DBLP:conf/eccv/KangRWRFV20, DBLP:journals/tnn/LuoZJLP23} aims to enhance functionality by enabling font style transfer.
Considering the powerful generative capabilities of diffusion models, some recent methods~\cite{DBLP:conf/icdar/NikolaidouRCSSSML23, DBLP:conf/aaai/YangPKZYJ24, DBLP:conf/cvpr/ZhuLWHY23} have begun to utilize diffusion models for local text generation, and achieving impressive results.
These methods achieve accurate generation of complex characters and show remarkable performance in font style transfer.
However, these methods often generate text on a blank canvas, ignoring the background of the text, which makes it difficult to apply them in natural scenes.
Image editing is the task most relevant to our task. Some methods~\cite{DBLP:conf/cvpr/AvrahamiLF22, DBLP:conf/cvpr/BrooksHE23, DBLP:conf/iclr/CouaironVSC23} based on diffusion models have achieved seamless object removal, replacement, and addition in images. 
However, these methods are not specifically designed for text and encounter significant difficulties when generating text.
In this paper, we achieve seamless text generation through local text generation based on diffusion models.

\section{Proposed method}
\label{sec3}
\begin{figure*}[t]
  \centering
  \includegraphics[width=0.7\linewidth]{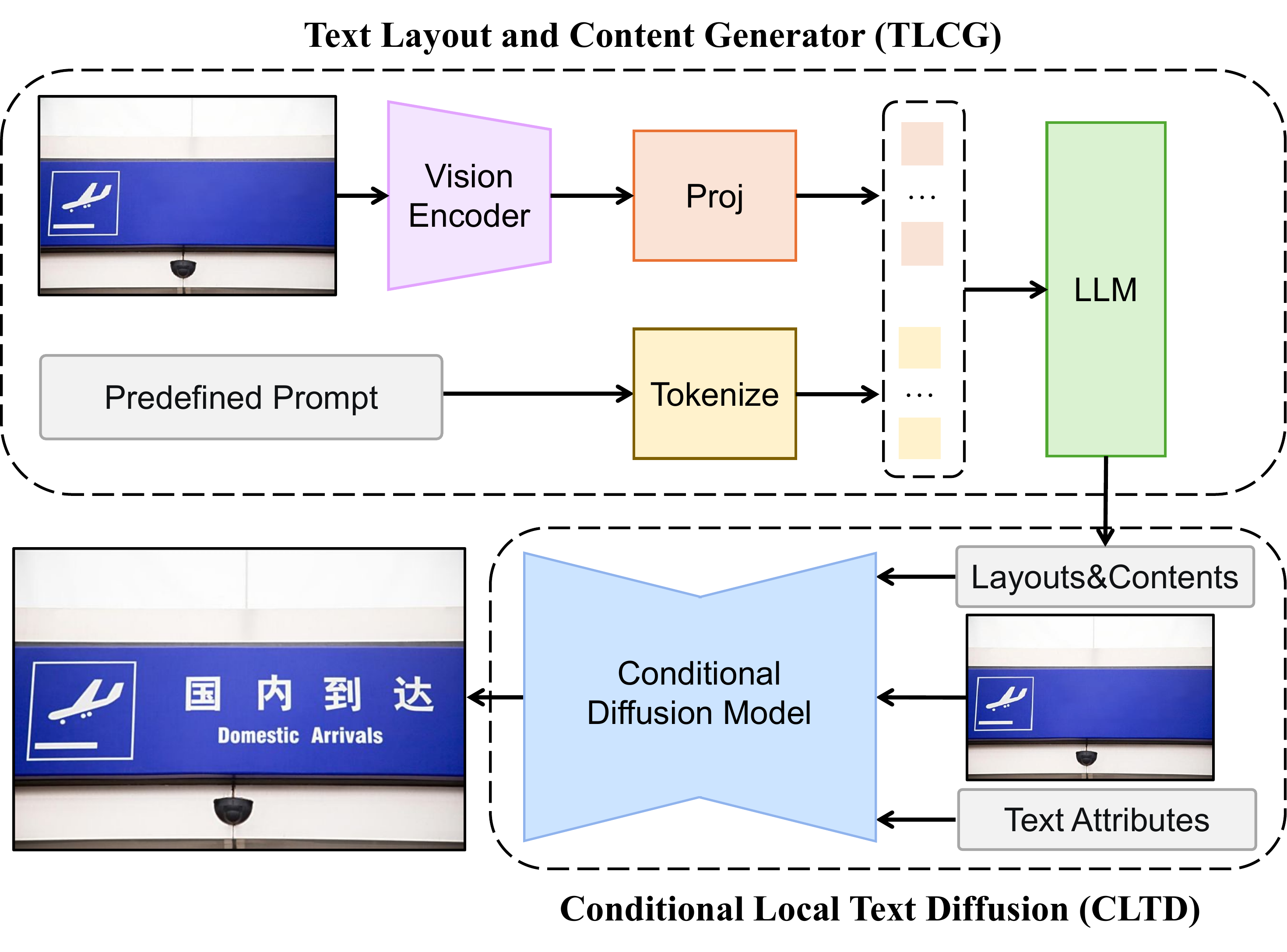}
  \vspace{-3mm}
  \caption{The overall pipeline of SceneVTG++. 
  With background images and predefined text prompts as input, TLCG generates reasonable text layouts and recommends appropriate text contents. CLTD then generates text on the background image based on TLCG outputs along with some other text attributes.}
  \label{fig:3}
\end{figure*}

The proposed model consists of two components: the Text Layout and Content Generator (TLCG) and the Controllable Local Text Diffusion (CLTD). We will describe the two parts in~\Cref{subsec3.1} and 3.2 respectively. \Cref{fig:3} shows the overall pipeline of SceneVTG++, in which TLCG generates text layouts and contents in nature languge and the CLTD generates text images based on various conditions. 
\subsection{Text Layout and Content Generator}
\label{subsec3.1}
Through large-scale pre-training with unsupervised data and fine-grained tuning with instruction-following data, MMLMs are capable of handling various visual reasoning tasks, such as open-set detection~\cite{wei2023lenna} and open-set segmentation~\cite{lai2024lisa}.
TLCG applies the instruction tuning method of large multimodal models to text layout and generation tasks.
Compared to previous work, TLCG stands out for harnessing the visual captioning capabilities of MMLMs to generate reasonable text layouts and compelling text contents in natural scenes.
TLCG is based on the pre-trained LLaVA~\cite{liu2024visual} and interprets tasks through full fine-tuning of LLM, as well as fine-tuning the visual encoder and projection layer to achieve task-specific visual feature extraction and transformation.

\begin{figure*}[t]
  \centering
  \includegraphics[width=0.98\linewidth]{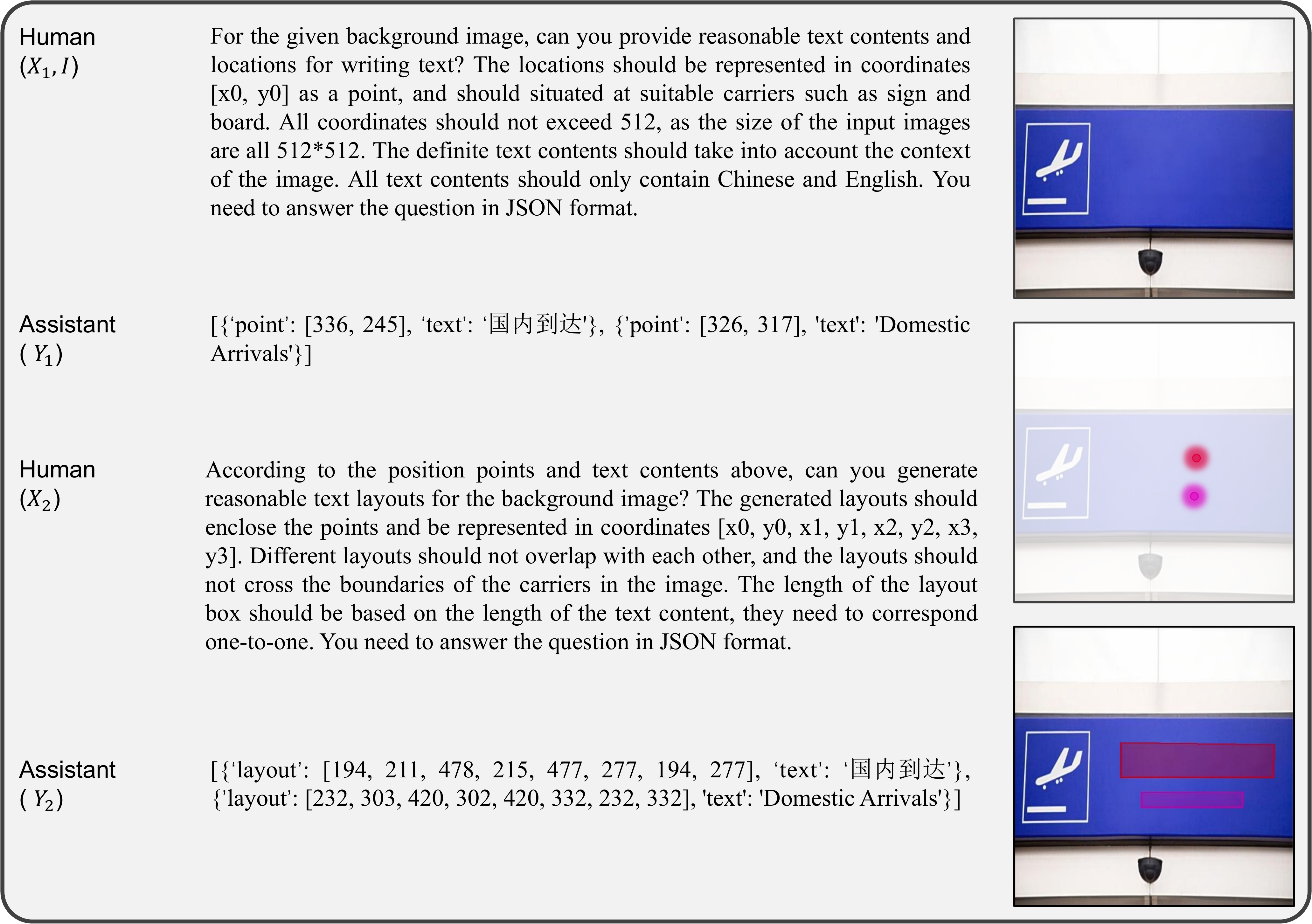}
  \caption{An example of TLCG workflow that generate reasonable text layout and content in two steps. The initial step involves identifying proposal points and contents of the text, followed by generating the text box in the next step.}
  \label{fig:4}
\end{figure*}

Drawing inspiration from the remarkable effectiveness of instruction-following fine-tuning methods~\cite{liu2024visual, liu2024improved, bai2023qwen, DBLP:conf/nips/Dai0LTZW0FH23, DBLP:journals/corr/abs-2308-10792}, we design a novel two-step prompt template to fine-tune the model for representing all dialogues. An example is shown in~\Cref{fig:4}:
(i) In the first step, given the first prompt $X_1$ and the background image $I$, TLCG locates reasonable proposal points for writing text and recommends semantically reasonable text contents for each points. The answer $Y_1$ is in JSON format, with each item containing two parts: point coordinates and text content string.
(ii) In the second step, according to the proposal points and the text contents, given the prompt X2, TLCG determines the layout position suitable for generating the text. The answer $Y_2$ is also in JSON format and consists of layout boxes and text contents.
We believe that the two-step prompt template has the following two main advantages: (1) chain-of-thought (cot) is a method to reduce hallucinations in large language models. The two-step template is similar to it because it breaks down a difficult task into two simpler tasks. (2) The length of the text layout boxes can be determined based on the length of the text contents, reducing the mismatch issues.
In addition, leveraging MMLMs can also manage various layout generation tasks by simply modifying the input prompt. 
For example, the language of the generated text can be specified in $X_1$ to adapt to more application scenarios. 
Considering the semantic coherence of the text contents and the reasonability of text layouts arrangement, TLCG generates text content taking text lines as basic units.

\subsection{Controllable Local Text Diffusion}
\label{subsec3.2}
\begin{figure*}[t]
  \centering
  \includegraphics[width=0.98\linewidth]{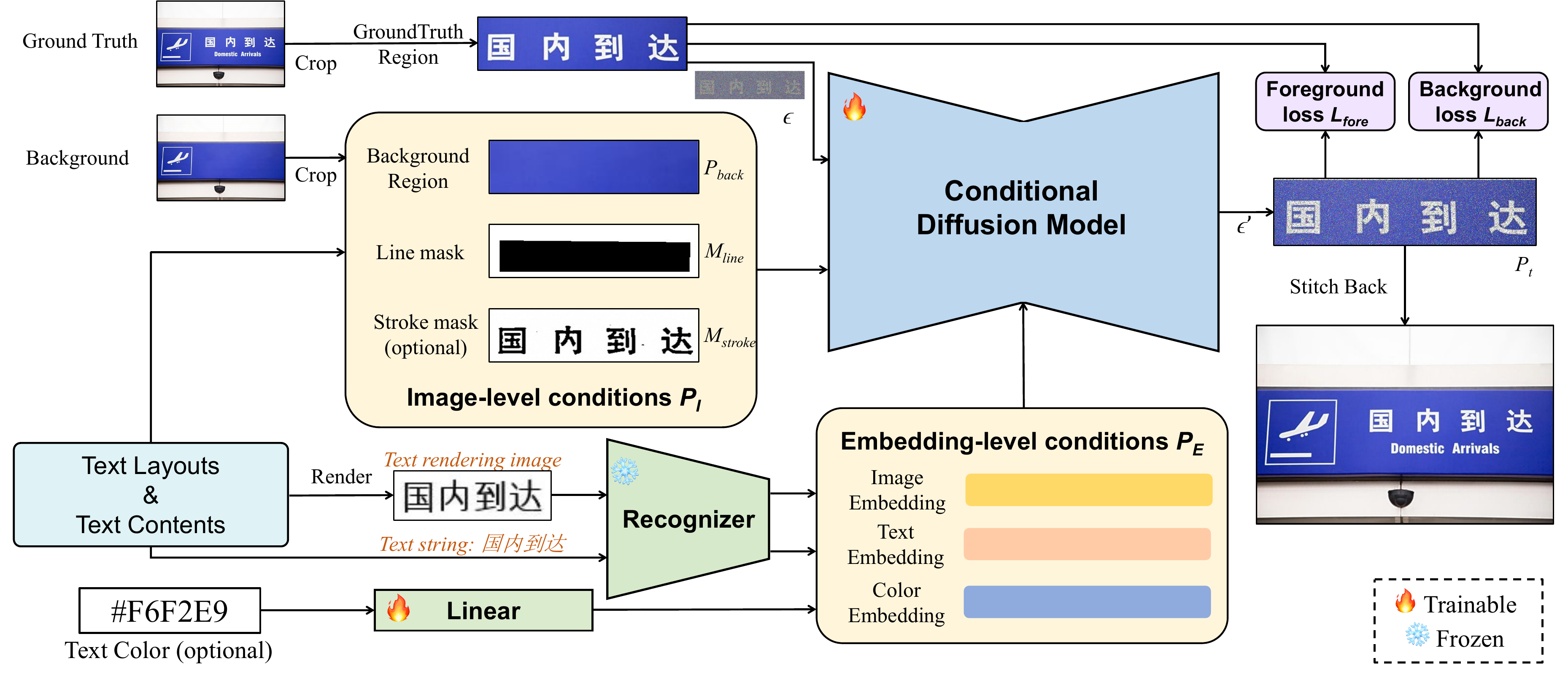}
  \caption{The detailed architecture of the Controllable Local Text Diffusion (CLTD). Taking the text layouts, contents, attributes and background images as input, CLTD renders desired text upon the background image.}
  \label{fig:5}
\end{figure*}

As described in~\Cref{subsec2.2.2}, both the full image generation and font generation have certain limitations in natural scenes. 
Therefore, the CLTD we propose focuses on generating text while also considering the integration of the text with the background image.
To align with the generation format of the TLCG phase, CLTD takes text lines as the basic unit for generating text. 
\Cref{fig:5} illustrates the process of generating text images using CLTD, which primarily consists of three parts: image-level condition, embedding-level condition, and conditional diffusion model.

\subsubsection{Image-level Condition}
The image-level condition comprises three parts: the background image region $P_{back}$, the text line mask $M_{line}$, and the stroke mask $M_{stroke}$. $M_{stroke}$ is optional during inference, while $P_{back}$ and $M_{line}$ are generated automatically based on text layouts.

\noindent\textbf{Background Image Region.} According to the bounding boxes of the text layouts, the corresponding position of the background image is cropped to obtain $P_{back}$, providing background pixel information. 
During this process, the boundary is extended by 10\% to remain the extended non-text area. 
The expansion of the boundaries ensures that the model retains the surrounding background pixels while generating text, thus ensuring that there are no visible seams when the generated image is stitched back into the original image.

\noindent\textbf{Text Line Mask.} $M_{line}$ is a mask image generated for the text area, with the masked part representing the position of the text boxes, used to indicate the generated text area. 
The existence of text line mask makes it possible to obtain accurate OCR annotations to increase the utility.

\noindent\textbf{Stroke Mask.} $M_{stroke}$ is a mask representing the font condition, which is obtained through adaptive threshold binarization during training.
It is easy to get a specified font rendering version during inference stage.
The introduction of $M_{stroke}$ enhances the controllability of generated text.

\subsubsection{Embedding-level Condition}
The Embedding-level condition is composed of image embedding, text embedding, and color embedding. 

\noindent\textbf{Image Embedding.} For image embedding, begin by employing a font rendering engine to generate a image of the specified text content.
Then, image embedding is extracted by a pre-trained OCR recognizer.

\noindent\textbf{Text Embedding.} Text embedding is obtained using the pre-trained classifier head of recognizer. 
Specifically, by inverting the structure and weights of the classification head, the original output layer is used as the input layer, allowing text tokens as input to obtain text embedding features.

\noindent\textbf{Color Embedding.} Color embedding is extracted using a trainable linear layer, with the input color represented in RGB tensor. 
During inference, the color condition is optional, and both image embedding and text embedding are extracted based on the text contents.

\subsubsection{Conditional Diffusion Model}
Conditional diffusion model generates images conditioned on image-level conditions and embedding-level conditions. 
The image-level condition is concatenated with the noise $\epsilon$ and fed into the diffusion model, while the embedding-level condition is input as an embedding into the key and value of the cross-attention layers in the U-Net. 
Referring to Composer~\cite{DBLP:conf/icml/HuangC0SZZ23}, we project color embedding and image embedding and add them to the timestep embedding. 
Additionally, to generate text of arbitrary size, the diffusion model in CLTD does not rely on the latent features of the image.
Pixel-level denoising ensures the preservation of small text features, avoiding the latent compression by VAE.

The training objective of diffusion models measures the difference between the predicted noise and the actual noise $\epsilon$, with commonly used loss functions including mean squared error:
\begin{align}
L_{cdm} = ||\epsilon - \epsilon_{\theta}(P_t,P_I,P_E,t)||^2_2,  
\end{align}
where $\epsilon_\theta$ represents the UNet for predicting noise.

For the generation of foreground text and background images, we design two auxiliary losses, foreground loss $L_{fore}$ and background loss $L_{back}$, to further enhance the performance of the model.
The foreground loss measures the difference between the text features in the predicted image $P'$ and the original image $P_0$, and is calculated as follows:
\begin{align}
L_{fore} = ||\mathbb{F}(M_{line}\times P_0) - \mathbb{F}(M_{line}\times P')||^2_2,  
\end{align}
where $\mathbb{F}$ represents a OCR recognizer to extract the feature from the images.
In contrast to foreground loss, background loss computes the variations in pixel values between the predicted image and the source image:
\begin{align}
L_{back} = ||(1-M_{line})\times P_0 - (1-M_{line})\times P'||^2_2.  
\end{align}
In conclusion, the final training loss consists of three parts in total:
\begin{align}
L = L_{cdm} + \lambda_f \times L_{fore} + \lambda_b \times L_{back}, 
\end{align}
where $\lambda_f$ and $\lambda_b$ are hyperparameters. 

\subsection{Dataset construction}
\label{subsec3.3}
\begin{figure*}[t]
  \centering
  \includegraphics[width=0.98\linewidth]{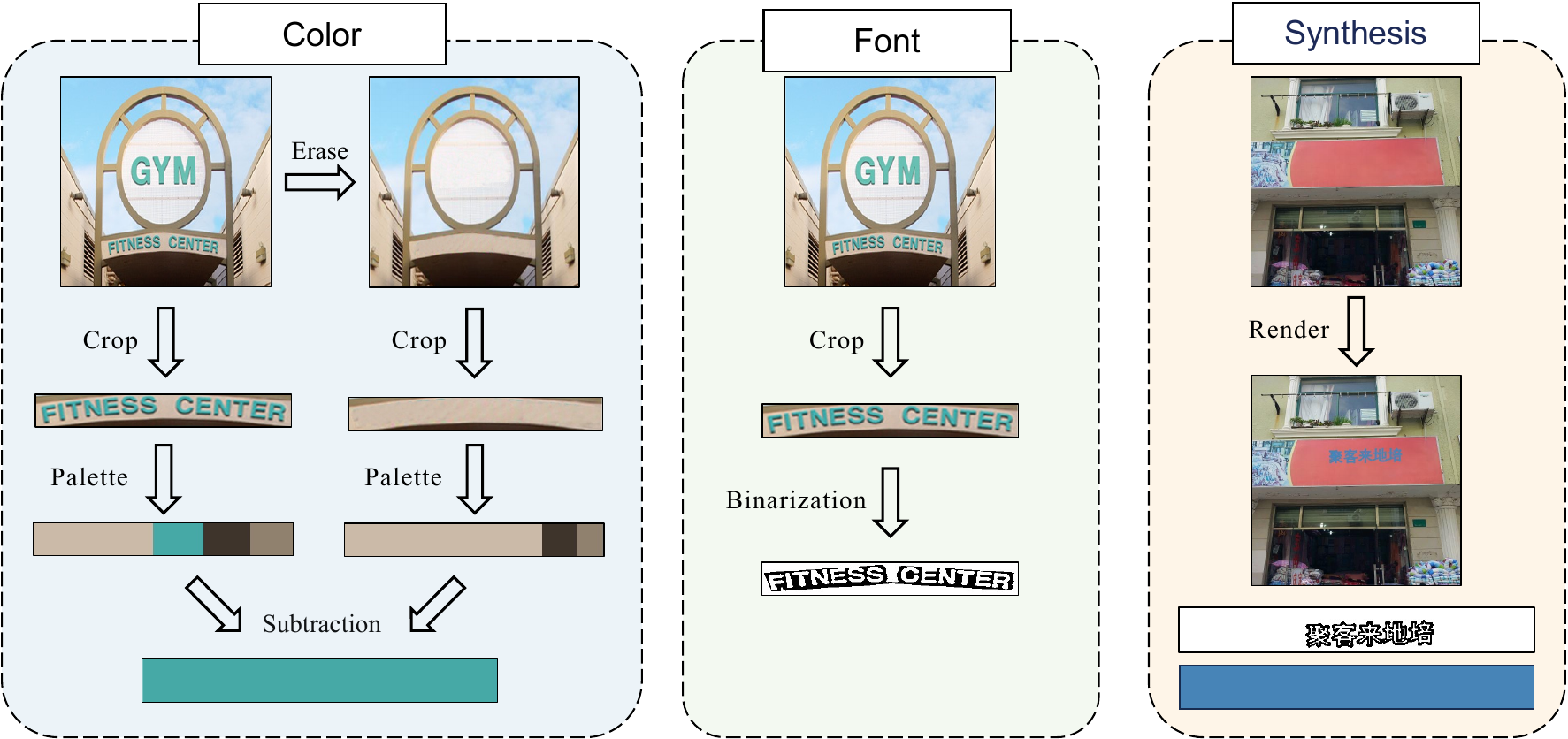}
  \caption{The process for constructing color and font conditions, as well as data synthesis.}
  \label{fig:6}
\end{figure*}

To achieve the controllability of color and font in generated text, we constructe color and font labels for the training data, as shown in~\Cref{fig:6}. 
For color condition, we first cropped the local regions containing text before and after text erasure. 
Then, we extracted the palettes of the regions before and after erasure through clustering. 
Finally, the color that decreased the most in the palette after erasure was considered the color of the text. 
For the construction of font labels, we cropped the text regions from the original image and extracted the font mask using adaptive threshold binarization. 
After that, we leverage a open-source OCR framework PaddleOCR~\cite{DBLP:journals/corr/abs-2206-03001} to filter out poorly segmented masks.
During inference, the stroke segmentation mask image is rendered based on the selected font to conditionally generate text. 
Due to the significant differences in the amount of different language data and the certain distribution bias of colors and fonts in nature scenes, we propose a synthetic dataset SceneVTG-Syn to expand the less languages data and to make the text colors and fonts more diverse.
The detailed statistics of the dataset will be described in~\Cref{secdata}

\section{Dataset}
\label{secdata}
The community has witnessed a significant increase in publicly available visual text data, ranging from MARIO~\cite{DBLP:conf/nips/ChenHL0CW23} to AnyWord~\cite{DBLP:conf/iclr/TuoXHGX24}.
However, when it comes to natural scene text erase data, the available amount is limited, and such datasets are crucial for training SceneVTG++. 
In this paper, we construct a real scene multilingual conditional text erase dataset SceneVTG-Erase++ and a synthetic dataset SceneVTG-Syn, and establishes a testset SceneVTG-benchmark++. 
The training data for SceneVTG-Erase++ comes from two sources, incorporating the publicly available scene text erase SCUTEnsText~\cite{DBLP:journals/tip/LiuLJZLW20} trainset. 
In addition, this paper collects multiple publicly available multilingual scene text datasets~\cite{DBLP:journals/corr/abs-1712-02170, DBLP:conf/icdar/KaratzasSUIBMMMAH13, DBLP:conf/icdar/KaratzasGNGBIMN15, DBLP:conf/cvpr/SinghPT0GH21, DBLP:journals/ijdar/ChngCL20, DBLP:conf/icdar/LongQPBFR23, zhang2017uber, DBLP:conf/icdar/ChngDLKCJLSNLNF19, DBLP:conf/icdar/ShiYLYXCBLB17, DBLP:conf/icdar/ZhangYBSKLJZJSL19, DBLP:conf/icdar/SunKCJNCLLNHDL19, DBLP:conf/icdar/NayefYBCFKLPRCK17, DBLP:conf/icdar/NayefLOPBCKKM0B19} and applies a scene text removal method CTRNet~\cite{DBLP:conf/eccv/LiuJLLCGD22} to generate erased text backgrounds from these datasets. 
After erasing, DiffBIR~\cite{DBLP:journals/corr/abs-2308-15070} is used to restore the images and smooth out any artificial traces in the erased areas. 
The dataset is an extension of the conference version dataset SceneVTG-Erase~\cite{zhu2024visual}, with further screening of the data, the introduction of more languages, and the construction of labels for color and font as mentioned in~\Cref{subsec3.3}.
Finally, SceneVTG-Erase++ contains approximately 197K images and a total of 1M text lines. 
In addition, we construct a 10K dataset for the evaluation of utility as~\Cref{subsec4.4} using COCOText~\cite{DBLP:journals/corr/VeitMNMB16} dataset in the same manner.
For the testset SceneVTG-benchmark++, in addition to the high-quality manually labeled SCUT-EnsText (annotated using Photoshop) testset, we also introduce the validation set of the multilingual natural scene dataset MLT2017~\cite{DBLP:conf/icdar/NayefYBCFKLPRCK17} and manually erase the text by Photoshop.

\section{Experiment}
\label{sec4}
In this section, we start with a detailed description of the experimental settings.
Then, we evaluate our framework from four aspects: fidelity, reasonability, utility, and controllability. 
Unless otherwise specified, all experiments below are evaluated on the SceneVTG-benchmark++.
\subsection{Implementation Details}
\label{subsec4.1}
Text Layout and Content Generator (TLCG) was fine-tuned based on the pre-trained LLaVA-v1.5-13b~\cite{liu2024improved}. 
When fine-tuning the model, we simultaneously enabled the training of the visual encoder, projection layer, and the LLM.
The learning rate for training these three modules was set to 2e-5.
We train TLCG for a total of 20 epochs on the SceneVTG-Erase++ dataset with a batch size of 16.
Controllable Local Text Diffusion (CLTD) refers to the Decoder of DALLE-2 but does not use pre-trained weights. 
We train CLTD from scratch for a total of 20 epochs on the SceneVTG-Erase++ dataset with a batch size of 128.
The text recognizer uses the CRNN~\cite{DBLP:journals/pami/ShiBY17} structure and is trained in natural scenes. 

\subsection{Experimental results of fidelity}
\label{subsec4.2}
We evaluate the fidelity of images generated by existing competing methods using the multilingual SceneVTG-benchmark++ constructed in~\Cref{subsec3.3} and compare English-only methods in the subset SceneVTG-Benchmark~\cite{zhu2024visual}. 
The methods for comparison include early rendering engine-based approaches like SynthText~\cite{DBLP:conf/cvpr/GuptaVZ16}, and the most recent state-of-the-art text generation methods based on diffusion models~\cite{DBLP:conf/iclr/TuoXHGX24, DBLP:journals/corr/abs-2311-16465}.
Following previous works, the evaluation metrics include Frechet Inception Distance (FID)~\cite{DBLP:conf/nips/HeuselRUNH17}, as well as OCR detection metric F-score (F) and OCR recognition metric Line Accuracy (LA). 
We use the open-source method PaddleOCR~\cite{DBLP:journals/corr/abs-2206-03001} as the model for calculating metrics.
The tasks evaluated in this section include end-to-end image generation and local text generation, designed to highlight the superiority of our method from multiple perspectives. 
For the local text generation task, we introduce a local FID metric that performs separate FID calculations on cropped areas containing text, thereby eliminating the interference from background images. 
The quantitative results are shown in~\Cref{tab:tab1}.

\begin{table}[t]
\caption{
Comparison of the fidelity on SceneVTG-benchmark++.
}
    \centering
    \vspace{1mm}
   \resizebox{0.98\textwidth}{!}
     {
    \begin{tabular}{ccccccccc}
    \toprule
    Language For & \multirow{2}{*}{Methods} &\multicolumn{3}{c}{End-to-end Inference}  &\multicolumn{4}{c}{Local Text Generation} \\
    \cmidrule(lr){3-5}\cmidrule(lr){6-9}
    Evaluation && FID$\downarrow$ & F$\uparrow$ & LA$\uparrow$   & FID$\downarrow$ & FID-R$\downarrow$ & F$\uparrow$ &LA$\uparrow$ \\
    \midrule   
    \multirow{7}{*}{English-Only} & SynthText~\cite{DBLP:conf/cvpr/GuptaVZ16} &    48.19 &       42.87 & 55.24 & 37.36   & 46.40 & 66.07 & 40.30\\
    & TextDiffuser~\cite{DBLP:conf/nips/ChenHL0CW23} &    78.35      &53.43  & 28.58 & -& -& -& -\\
    & GlyphControl~\cite{DBLP:conf/nips/YangGYLDH023} & 77.95     &  49.23 & 26.31& -& -& -&-\\
    &AnyText~\cite{DBLP:conf/iclr/TuoXHGX24}  & 73.07     &  44.77 & 37.17& 35.76&	48.03&	62.65&	12.84\\
    &TextDiffuser-2~\cite{DBLP:journals/corr/abs-2311-16465}  & 88.02    &  33.61 & 33.04& 42.00    & 47.93 & 69.26 & 26.13\\
    & SceneVTG~\cite{zhu2024visual} &  \textbf{26.28}& 52.03 & \textbf{75.62}& 27.66 & 33.34 & \textbf{75.73} & \textbf{53.21}\\
    & SceneVTG++ &  29.99&	\textbf{58.71}&	74.93&\textbf{26.22}	&\textbf{30.94} &71.94	&49.61\\
    \midrule   
    \multirow{2}{*}{Multi-Language} & AnyText~\cite{DBLP:conf/iclr/TuoXHGX24} &81.45	&40.23&	24.59 & 38.13&	41.31&	\textbf{55.05}&	10.59\\
    & SceneVTG++&\textbf{31.55}&	\textbf{58.31}	&\textbf{62.38} &\textbf{26.83}	&\textbf{18.21}	&50.50	&\textbf{42.71}\\
      \bottomrule
\end{tabular}
     } 
    \label{tab:tab1}
\end{table}


\subsubsection{End-to-end Inference} 
For end-to-end image generation tasks, we make our best effort to ensure that different generative paradigm methods are compared in an aligned manner.
Specifically, for the rendering engine-based method SynthText, SceneVTG~\cite{zhu2024visual} and the proposed SceneVTG++, we use the entire process of the method, starting from a background image without text, to generate text for the image.
For diffusion-based text generation methods, since they cannot determine the text layout themselves, they generate text according to the same layout and determine the background of the image based on the caption generated by BLIP-2~\cite{li2023blip}.
As can be seen from~\Cref{tab:tab1}, for English image generation, SceneVTG++ exhibits competitive results compared to SceneVTG and significantly outperforms other previous methods in terms of FID, F, and LA.
It is worth noting that SceneVTG can only generate images with English text. 
In the task of end-to-end generation of images with multilingual text, currently only AnyText~\cite{DBLP:conf/iclr/TuoXHGX24} has achieved the generation of multilingual text, but its performance is far inferior to that of proposed SceneVTG++ in all three metrics.
\begin{figure*}[t]
  \centering
  \includegraphics[width=0.98\linewidth]{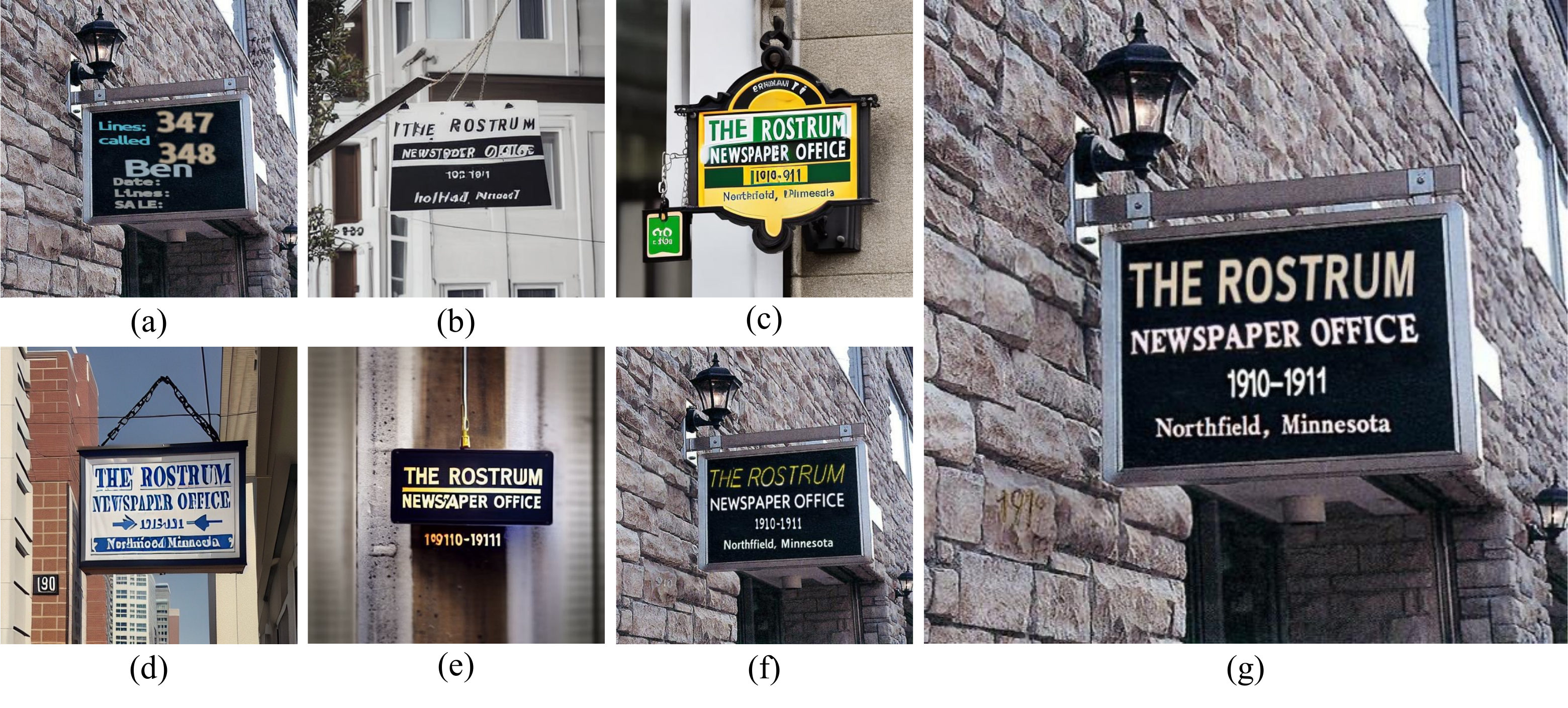}
  \vspace{-2mm}
  \caption{
  Qualitative results of end-to-end English text generation.
  (a) SynthText; (b) Textdiffuser; (c) GlyphControl; (d) AnyText; (e) Textdiffuser-2; (f) SceneVTG; (g) SceneVTG++. Among them, SynthText, SceneVTG, and SceneVTG++ generate text on given background images, while the other methods generate images based on the same layout and caption. Zoom in for better views.}
  \label{fig:7}
\end{figure*}

\begin{figure*}[thb]
  \centering
  \includegraphics[width=0.98\linewidth]{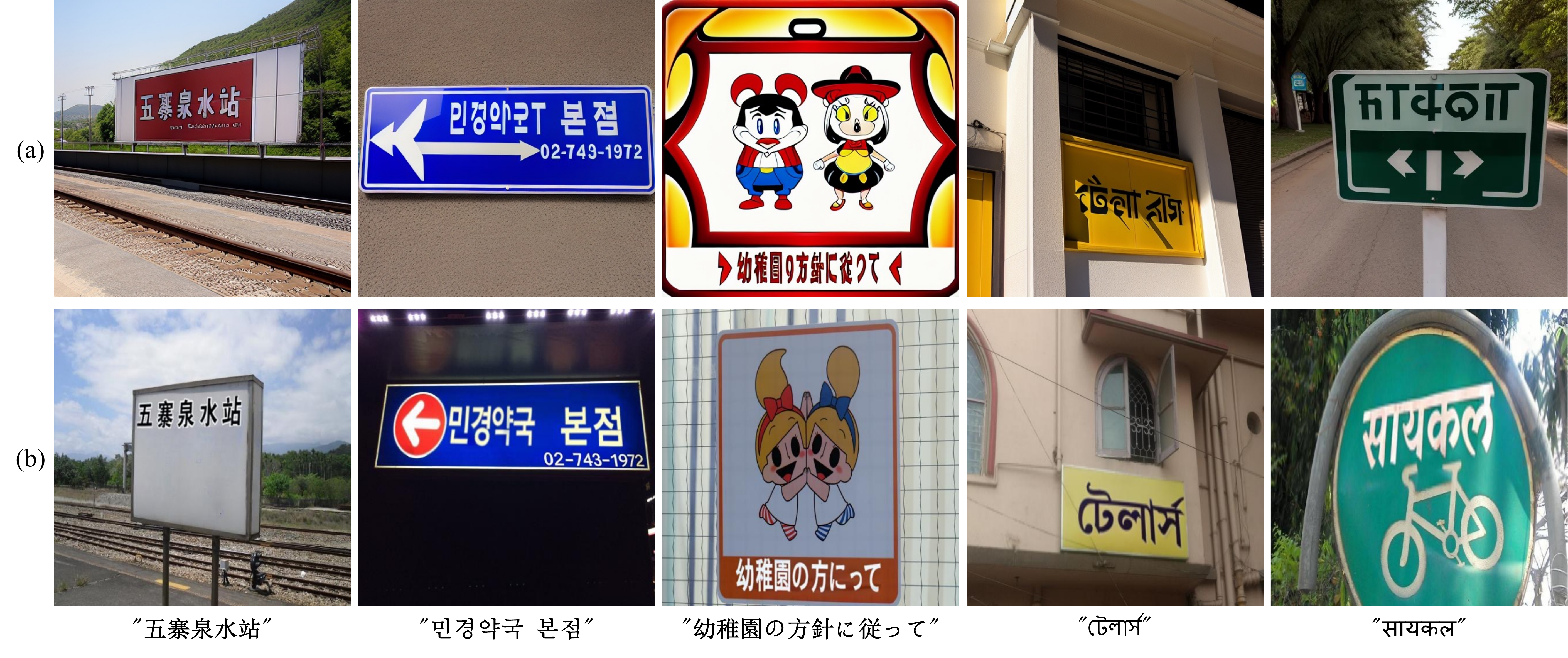}
  \vspace{-2mm}
  \caption{Visualizations of end-to-end multilingual text generation results: (a) AnyText; (b) SceneVTG++. The texts generated from left to right are: (i) ``Wuzhai Spring Water Station''; (ii) ``Minkyung Pharmacy Main Branch''; (iii) ``In accordance with the policy of the kindergarten''; (iv) ``Tailors''; (v) ``Bicycle''. Zoom in for better views.}
  \label{fig:8}
\end{figure*}

\Cref{fig:7} shows the qualitative results in end-to-end generation of English text. 
It can be seen that the four diffusion model-based methods ((b), (c), (d), (e)) are more likely to produce characters with stroke errors, especially when the generated text size is small.
This also results in poorer recognition metrics (LA) for these four methods as shown in~\Cref{tab:tab1}.
The rendering-based method (a) can generate accurate text, but it blends poorly with the background, showing obvious artificial traces. 
The layout of the generated text is also unreasonable, and a discussion about this will be presented in the next section. 
Method (f) is the conference version of our method, focusing on English text generation, and it exhibits good fidelity.
\Cref{fig:8} shows the qualitative results of multilingual text generation, 
and we can see that SceneVTG++ still performs better than Anytext in generating multilingual texts.

\subsubsection{Local Image Generation} 
In addition to end-to-end image generation, we also explored the performance of different methods in local text generation tasks in detail, to more comprehensively evaluate the fidelity of the generated text. 
For this task, all methods adopted the same inference approach: generating specified text at designated layouts on a background image.
The results in~\Cref{tab:tab1} show that SceneVTG++ exhibits competitive performance on the English text generation task. 
Moreover, for the multilingual local text generation task, SceneVTG++ significantly outperforms previous methods across all four metrics.
As not all methods support inpainting mode, we only compared some of the methods for this task. 

\begin{figure*}[thb]
  \centering
  \includegraphics[width=0.98\linewidth]{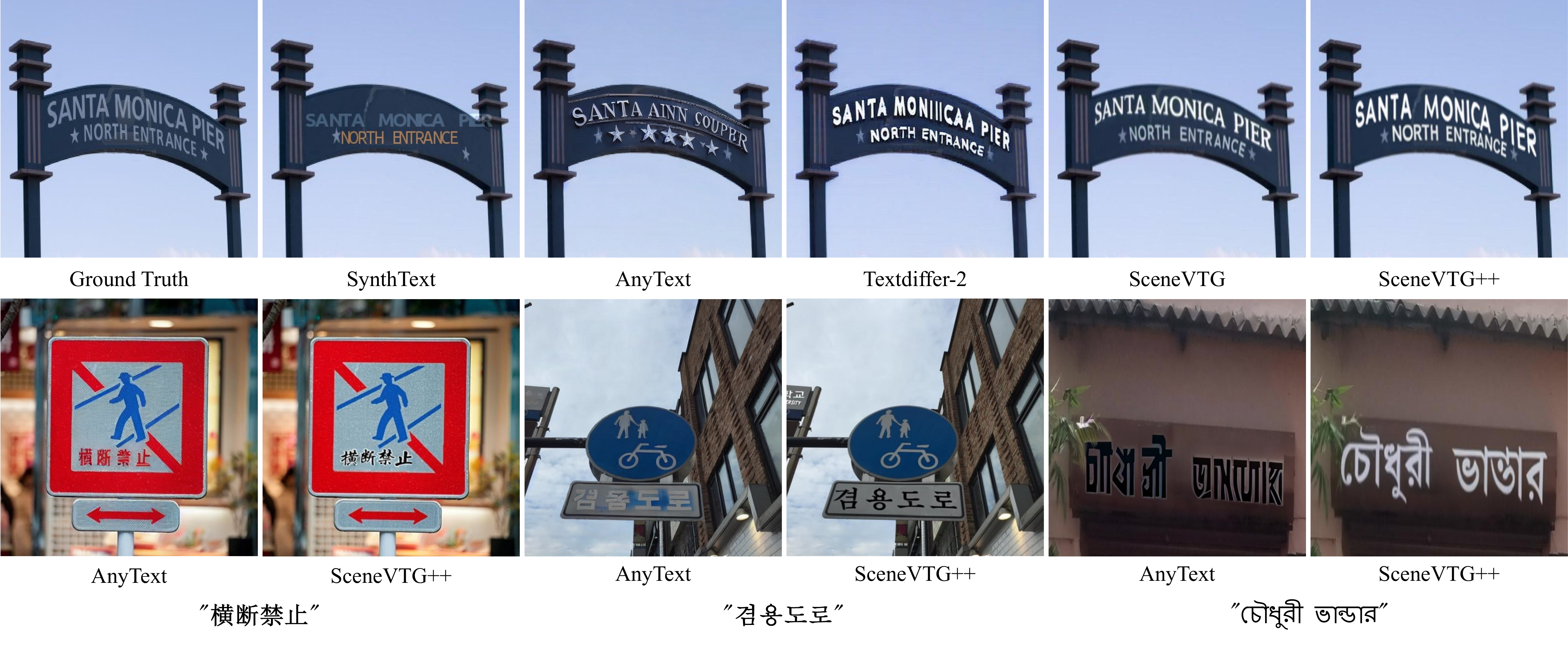}
  \vspace{-2mm}
  \caption{Visualizations of local text generation results. The first row visualizes local English text generation results between SceneVTG++ and other methods, while the second row contrasts SceneVTG++ with AnyText in multi-lingual local text generation. Zoom in for better views.}
  \label{fig:9}
\end{figure*}

\Cref{fig:9} shows the visualization of local text generation for different methods in English and multilingual text. 
In English generation, the text quality generated by SceneVTG++ is superior to that of AnyText and TextDiffuser-2. 
While SynthText produces clear text, it does not blend well with the background. 
For multilingual local text generation, SceneVTG++ can generate more accurate characters compared to AnyText.


\subsection{Experimental results of reasonability}
\label{subsec4.3}

The evaluation metrics for reasonability include three layout-related indicators: IOU, PD-Edge, and Readability score, as well as one content-related metric: CLIPScore. 
For IOU, since the layout generation task is open-ended, it is not possible to have a one-to-one correspondence between predicted boxes and ground truth boxes. 
Therefore, we calculate the IoU between all predicted boxes and ground truth boxes without requiring a one-to-one match. 
For the PD-Edge metric, we propose to use PiDiNet~\cite{DBLP:conf/iccv/0002LYH00P021} for edge detection.
To be specific, PiDiNet is used to detect the edges of the background image for computing the total pixel of the edge regions crossed by the generated layout. 
The rationale behind this metric is that, in the real world, text rarely appears on prominent boundaries. 
The Readability score is based on previous work~\cite{seol2024posterllama}, evaluating the clarity of the generated text according to the gradient changes of the text in the image space.
For CLIPScore, we utilize a pre-trained CLIP~\cite{DBLP:conf/icml/RadfordKHRGASAM21} model to assess the relevance between the image and the generated text.
In this section, we compare the segmentation network-based method SynthText and the large language model-based method PosterLlama~\cite{seol2024posterllama}.

\begin{figure*}[t]
  \centering
  \includegraphics[width=0.98\linewidth]{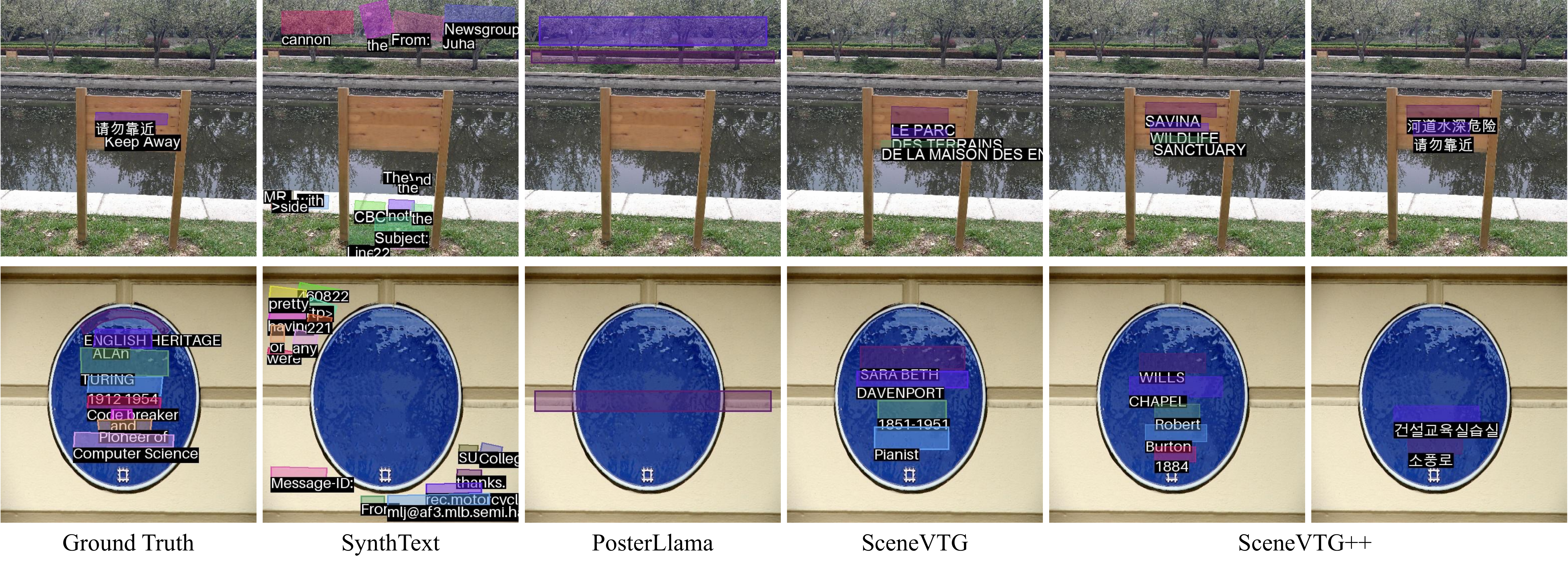}
  \vspace{-2mm}
  \caption{Visualizations of text layout and content generation. (a) and (b) show the comparison between SceneVTG++ and other methods, demonstrating that the generated layouts and content are more reasonable. 
  (c) demonstrates the capability of SceneVTG++ to generate multilingual text content, with the generated text from top to bottom in the right column: (i) ``Water depth dangerous do not approach''; (ii) ``Construction Training Room of Picnic Road''. Zoom in for better views.}
  \label{fig:10}
\end{figure*}

\begin{table}[thb]
\caption{Comparison of text layout and content generation reasonability with multiple existing methods.}
\vspace{1mm}
\centering
\resizebox{0.75\textwidth}{!}
     {
    \centering
    \begin{tabular}{ccccc}
    \toprule
Method     & IoU$\uparrow$  & PD-Edge$\downarrow$ & Readability$\downarrow$ & CLIPScore$\uparrow$ \\
    \midrule
    SynthText~\cite{DBLP:conf/cvpr/GuptaVZ16} & 7.31 & 477.12    &29.32  & 21.74 \\
    PosterLlama~\cite{seol2024posterllama}	& 5.00 & 2261.52		&28.81&-\\
    SceneVTG~\cite{zhu2024visual}&	\textbf{33.95} 	&752.94		&25.19&22.59\\
SceneVTG++&	33.52	&\textbf{405.28}		&\textbf{25.03}&\textbf{22.73}\\
    
\bottomrule
    \end{tabular} 
    }
    \label{tab:tab2}
\end{table}

\Cref{tab:tab2} presents the quantitative experimental results, which shows that SceneVTG++ achieves the best performance across multiple metrics, indicating that the generated layouts are the most reasonable.
It is worth noting that during the layout and content generation phase, SceneVTG++ is capable of generating content in multiple languages and focuses on generating quadrilateral text boxes, which is the most significant difference from the conference version~\cite{zhu2024visual}.
\Cref{fig:10} shows a comparison of some qualitative results. 
From (a) and (b), it can be seen that compared to SynthText and PosterLlama, our method generates more reasonable text layouts and content. 
(c) demonstrates the capability of SceneVTG++ to generate reasonable multilingual text.
\begin{table}[h]
 \caption{
 Comparison of the utility on SceneVTG-benchmark++ by training text detectors and recognizers using synthetic data.
 ``Eng-Re'' denotes the average recognition results of English regular datasets, including IIIT, SVT, and IC13 benchmark.``Eng-Ir'' denotes the results of English irregular datasets including IC15, SVTP, and CUTE benchmark. ``Mul'' denotes the results on multilingual benchmark MLT2017.}
  \vspace{1mm}
    \centering
   \resizebox{0.85\textwidth}{!}
     {
    \begin{tabular}{cccccccc}
    \toprule
    \multirow{2}{*}{Methods}    &\multicolumn{3}{c}{Detection Results~(F$\uparrow$)}  &\multicolumn{4}{c}{Recognition Result~(LA$\uparrow$)} \\
    \cmidrule(lr){2-4}\cmidrule(lr){5-8}
     &  IC13 & IC15 & MLT17  & Eng-Re & Eng-Ir & Eng-Avg & Mul\\
    \midrule    
        SynthText &    72.38&      56.65&      47.17  &48.55 & 26.08 & 39.74&25.65\\
        VISD &    75.19&     65.49&     50.21  & 38.02 & 26.88 & 33.65& - \\
        UnrealText  &    73.73&     61.80&    47.62 & 32.06 & 19.85 & 27.28 & - \\
        LBTS  &    68.83&     52.69&     40.59 & -  & - & - & - \\
    \midrule    
        TextDiffuser  &    52.50&     34.08&     29.06& 30.08 & 8.79 & 21.74 & - \\
        GlyphControl  &    35.04&    40.64&     24.37 & 40.80 & 13.21 & 29.99& - \\
        AnyText  &  48.05 &    42.45&    30.75 & 49.80 & 19.62 & 37.97&5.35\\
        TextDiffuser-2  &    21.47&      10.79&     12.23  & 40.26 & 10.48 & 28.59& - \\
    \midrule    
        Real data  & \underline{76.34} & \underline{79.15} & \underline{56.09}  & 57.97 & \underline{37.66} & \underline{50.01}& \underline{52.51} \\
        SceneVTG   &    \textbf{75.36}&     \textbf{66.31}&     \textbf{53.90} & 54.97 & 35.50 & 47.34& - \\
        SceneVTG++&   74.68 &     65.13&    53.51 & \textbf{58.21}	&\textbf{36.03}	&\textbf{49.52}  &\textbf{38.28}\\
    \bottomrule
    \end{tabular} 
     } 
    \label{tab:tab3}
\end{table}

\begin{figure*}[!h]
  \centering
  \includegraphics[width=0.98\linewidth]{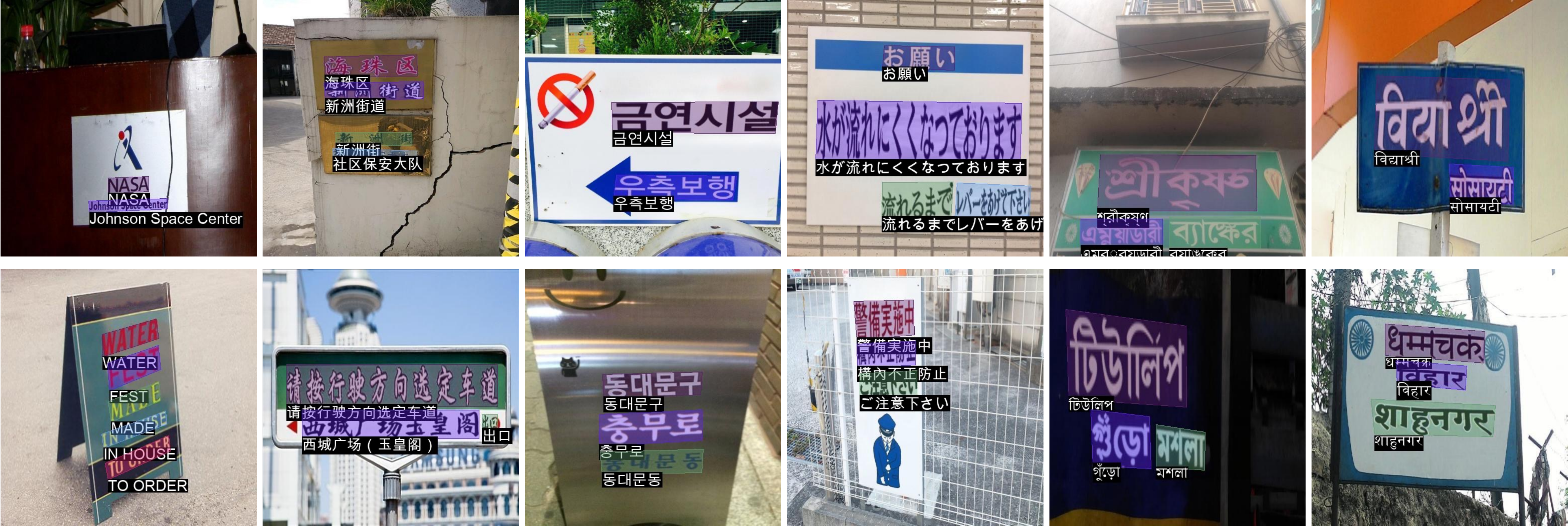}
  \vspace{-2mm}
  \caption{Visualizations of generated text with annotations. SceneVTG++ can generate text at the given location with accurate characters, ensuring the accuracy of annotations when training OCR tasks. Zoom in for better views.}
  \label{fig:11}
\end{figure*}

\subsection{Experimental results of utility}
\label{subsec4.4}
\begin{figure*}[th]
  \centering
  \includegraphics[width=0.98\linewidth]{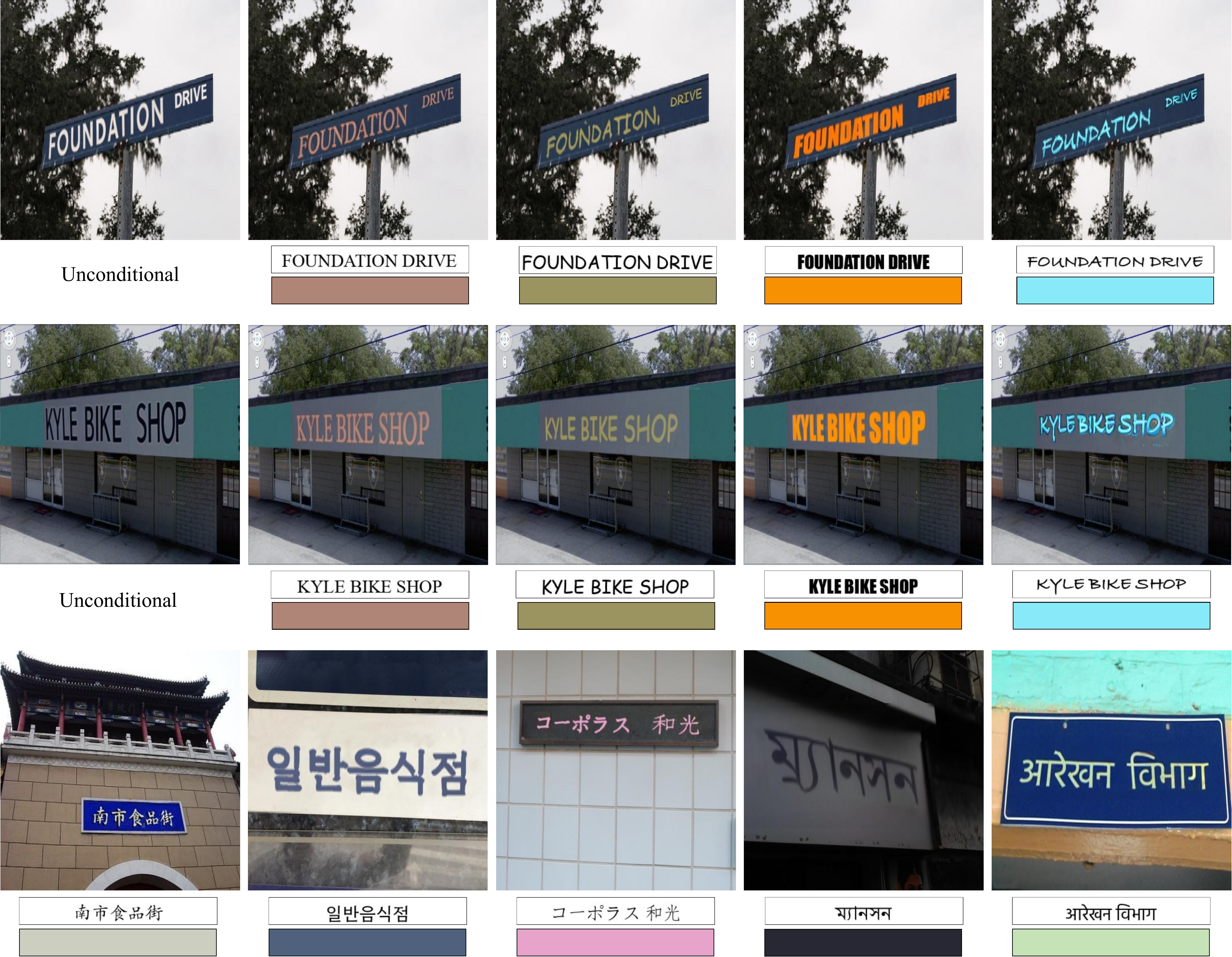}
  \vspace{-2mm}
  \caption{Visualizations of generated text with attribute conditions. SceneVTG++ is capable of achieving excellent control over text attributes, such as color and font control. Zoom in for better views.}
  \label{fig:12}
\end{figure*}
We evaluate the utility of the generated text images by training text detection task and text recognition task.
\Cref{tab:tab3} shows the results of training OCR tasks using the same amount of generated images on multiple benchmarks~\cite{DBLP:conf/icdar/KaratzasGNGBIMN15, DBLP:conf/icdar/KaratzasSUIBMMMAH13, DBLP:conf/icdar/NayefYBCFKLPRCK17, DBLP:conf/bmvc/MishraAJ12, DBLP:conf/iccv/WangBB11, DBLP:journals/eswa/RisnumawanSCT14}.
For methods based on rendering engines, we directly use the images provided in their papers for training. 
For diffusion model-based methods, we perform end-to-end image generation on the text-erased data built from COCOText as described in~\Cref{subsec3.3}.
\Cref{tab:tab3} shows the experimental results of utility, where SceneVTG++ demonstrates competitive performance in both detection and recognition tasks.
Particularly, SceneVTG++ has shown significant improvements in multilingual recognition tasks.
\Cref{fig:11} demonstrates the utility of the text generation by SceneVTG++. 
It can be seen that the generated text and its placement are well-aligned with the annotations.

\subsection{Experimental results of controllability}
\label{subsec4.5}

\Cref{fig:12} shows the effect of SceneVTG++ in controlling text attributes.
When no conditions are provided, SceneVTG++ generates text with random attributes based on the background image. 
Once font and color attributes are specified, the model generates text that follows the given attributes.
For color attributes, it is only necessary to input the RGB values of the desired color into the model to achieve control. 
For font attributes, rendering a font image using the desired font is the first step.
Then construct a stroke mask to control the generated characters.
It is worth noting that SceneVTG++ has implemented precise attribute control for other languages besides English, such as Chinese, Korean, and so on.

\begin{table}[thb]
\caption{Ablation study of TLCG. We compared two different strategies for training multimodal models.}
\vspace{1mm}
\centering
\resizebox{0.75\textwidth}{!}
     {
    \centering
    \begin{tabular}{ccccc}
    \toprule
Method     & IoU$\uparrow$  & PD-Edge$\downarrow$ & Readability$\downarrow$ & CLIPScore$\uparrow$ \\
    \midrule
    LoRA fine-tuning & 30.64&	420.49&	\textbf{24.65}& \textbf{24.75} \\
Full fine-tuning&	\textbf{33.52}	&\textbf{405.28}	&	25.03& 22.73\\

\bottomrule

    \end{tabular} 
    }
    \label{tab:tab4}
\end{table}

\subsection{Ablation Study}
\label{subsec4.6}

In this subsection, we conducted ablation studies on the components proposed in the paper. 
\Cref{tab:tab4} shows the results of the TLCG ablation experiments. 
As can be seen from the table that when using full fine-tuning to train the large multimodal model, the performance of the model is better than that of LoRA fine-tuning. 
In addition to the readability score, full fine-tuning performed better than LoRA fine-tuning on all other metrics.
This indicates that the changes in training strategies in this paper compared to the conference version are effective. 
\Cref{tab:tab5} presents the ablation study results for CLTD. 
From the table, it can be observed that the SceneVTG-Syn data constructed in this paper is crucial. 
Better LA score indicates that the text generated by the model trained on synthetic dataset SceneVTG-Syn is more accurate.
We contend that the primary reason is the uneven distribution of different languages and characters in the real training data, and introducing synthetic data plays a balancing role.

\begin{table}[hb]
\caption{Ablation study of CLTD. We compared the preformance of training the model with and without synthetic dataset SceneVTG-Syn.}
\vspace{2mm}
\centering
\resizebox{0.6\textwidth}{!}
     {
    \centering
    \begin{tabular}{ccccc}
    \toprule
SceneVTG-Syn     & FID$\downarrow$  & FID-R$\downarrow$ & F$\uparrow$ & LA$\uparrow$ \\
    \midrule
     &	27.32	&19.34	&\textbf{55.44}& 40.02\\
    $\checkmark$ & \textbf{26.83}& \textbf{18.21}& 50.50 &\textbf{42.71} \\

\bottomrule
    \end{tabular} 
    }
    \label{tab:tab5}
\end{table}
\vspace{-5mm}
\section{Conclusion}

\vspace{-2mm}
\label{sec5}
In this paper, we explore the challenges in existing natural scene text generation methods and introduce a framework called SceneVTG++.
Specifically, it consists of two stages: TLCG and CLTD. 
By leveraging multimodal large language model and diffusion model, SceneVTG++ achieves controllable multilingual text generation in natural scenes.
We validate the superiority of the proposed SceneVTG++ method from four aspects: fidelity, reasonability, utility, and controllability. 
In addition to the model, we also contribute a new real dataset, SceneVTG-Erase++, and a synthetic dataset, SceneVTG-Syn. 
We hope these datasets will provide resources for researchers and practitioners in the field
Our future research will focus on reducing inference costs and exploring the possibility of an end-to-end process.
\vspace{-3mm}
\section*{Acknowledgements}
\vspace{-2mm}
This work was supported by the National Science and Technology Major Project under Grant No. 2023YFF0905400 and the National Natural Science Foundation of
China (No. 62376102).

\newpage
\bibliographystyle{elsarticle-num-names.bst} 
\bibliography{egbib}
\end{document}